\documentclass[conference]{IEEEtran}
\IEEEoverridecommandlockouts

\usepackage{amsmath, amssymb, amsfonts}

\usepackage{algorithm}
\usepackage{algpseudocode}

\usepackage{graphicx}
\usepackage{xcolor}
\usepackage{textcomp}
\usepackage{float} % To force algorithm placement using [H]

\usepackage{cite}

\def\BibTeX{{\rm B\kern-.05em{\sc i\kern-.025em b}\kern-.08em
    T\kern-.1667em\lower.7ex\hbox{E}\kern-.125emX}}
    
\begin{document}

\title{Low-Rank Matrix Approximation for Neural Network Compression}

\author{\IEEEauthorblockN{Kalyan Cherukuri\textsuperscript{1*}, Aarav Lala\textsuperscript{1*}}  
\thanks{\textsuperscript{*} Equal contribution as first authors.}  
\thanks{\textsuperscript{1} Kalyan Cherukuri and Aarav Lala are with the Illinois Mathematics and Science Academy ({\tt kcherukuri, alala1})@imsa.edu.}  }
\vspace{-1.5cm}  % Adjust the negative space as needed

\maketitle

\begin{abstract}
Deep Neural Networks (DNNs) have encountered an emerging deployment challenge due to large and expensive memory and computation requirements. In this paper, we present a new Adaptive-Rank Singular Value Decomposition (ARSVD) method that approximates the optimal rank for compressing weight matrices in neural networks using spectral entropy. Unlike conventional SVD-based methods that apply a fixed-rank truncation across all layers, ARSVD uses an adaptive selection of the rank per layer through the entropy distribution of its singular values. This approach ensures that each layer will retain a certain amount of its informational content, thereby reducing redundancy. Our method enables efficient, layer-wise compression, yielding improved performance with reduced space and time complexity compared to static-rank reduction techniques.

\end{abstract}


\section{Introduction}
Deep Neural Networks (DNNs) have demonstrated success across a wide range of tasks, from computer vision to natural language processing tasks \cite{choudhary2020comprehensive}. However, the high memory and computational demands of these deep models render them impractical for deployment in resource-constrained environments. The large number of hyperparameters in DNNs often exceeds the millions, resulting in high storage requirements and large computational costs, limiting both scalability and real-world applicability \cite{qararyah2021computational}. 

To address these emerging challenges, model compression has become a pressing need and an essential area for future research \cite{li2023model}. A promising approach is to reduce the size of neural networks while attempting to preserve performance, with a low-rank matrix approximation. Through approximating weight matrices in DNNs with lower-rank decompositions, such as with Singular Value Decomposition commonly referred to as SVD, the number of parameters involved for the network's computation can be significantly reduced, leading to lower memory usage and faster inference times \cite{li2023model}. However, conventional methods that apply a fixed rank reduction towards all layers of a network may encounter suboptimal trade-offs with compression and accuracy.

In this paper, we introduce a novel Adaptive-Rank Singular Value Decomposition (ARSVD) method that dynamically adapts the rank of weight matrices for each layer. This method that we propose and validate experimentally provides the foundation for a novel approach to the classical neural network compression task. To evaluate the effectiveness of our proposed algorithm, we compare the performance of several pre-trained models ResNet18, VGG16, and InceptionV3 under standard fixed-rank SVD compression, and compare results to our proposed entropy-guided ARSVD compression. These models are then tested on benchmark datasets including MNIST, CIFAR-10, and CIFAR-100. Experimental results demonstrate that ARSVD consistently achieves better compression-performance trade-offs than traditional fixed-rank SVD, preserving accuracy while significantly reducing model complexity.

The adaptability of ARSVD paves the way for more computationally efficient but complex neural network architectures, particularly to expand its applicability to real-world scenarios. In this paper, we aim to contribute: an entropy-guided adaptive rank selection, performance-efficient compression, and a scalable application across different architectures.
\vspace{-0.12 cm}
\section{Related Works}
\subsection{Neural Networks}
\vspace{-0.12 cm}
Neural networks have been a cornerstone of modern deep learning. Early models such as the perceptron \cite{rosenblatt1958perceptron} could only solve linearly separable problems, but the introduction of the backpropagation algorithm \cite{rumelhart1986learning} enabled efficient training of multi-layer perceptrons, significantly improving their expressive power.

The advent of Convolutional Neural Networks (CNNs), particularly LeNet-5 \cite{lecun1998gradient}, introduced architectural innovations such as local receptive fields, weight sharing, and spatial sampling concepts that became standard in deep learning. A major breakthrough came with AlexNet \cite{krizhevsky2012imagenet}, which outperformed traditional methods on the ImageNet benchmark and demonstrated the feasibility of training deep networks at scale. It also introduced the use of rectified linear units (ReLU) to enable robust non-linearity in deep networks.

More recently, the Transformer architecture introduced in the landmark \emph{Attention Is All You Need} \cite{vaswani2017attention} has shifted the paradigm away from convolutional and recurrent models by using self-attention mechanisms to capture global dependencies. This innovation has led to state-of-the-art performance across natural language and vision tasks. Vision Transformer (ViT) \cite{dosovitskiy2021an} further adapts this architecture to image classification, showcasing the flexibility and scalability of attention-based models in visual based tasks.

\vspace{-0.1 cm}
\subsection{Space-Time Complexity Benchmarks}

\begin{table*}[t]
\centering
\caption{Comparison of SVD-Based Compression Methods}
\label{tab:svd_comparison}
\scalebox{1}{%
\begin{tabular}{|l|l|l|l|}
\hline
\textbf{Method} & \textbf{Complexity} & \textbf{Performance} & \textbf{Limitations} \\
\hline
Truncated SVD (TSVD)  & 
\begin{tabular}{@{}l@{}}
Time: $O(kmn)$ \\ 
Space: $O(k(m+n))$
\end{tabular} & 
Accuracy: 5--15\% drop & 
Fixed-rank, ignores layer sensitivity \\
\hline
Randomized SVD (RSVD)  & 
\begin{tabular}{@{}l@{}}
Time: $O(mn\log k + k^2(m+n))$ \\
Space: $O(k(m+n))$
\end{tabular} & 
Accuracy: 3--10\% drop & 
Requires oversampling tuning \\
\hline
Transpose Trick  & 
\begin{tabular}{@{}l@{}}
Time: $O(m^3)$ when $m \ll n$ \\
Space: No reduction
\end{tabular} & 
Accuracy: None & 
Only applicable when $m \ll n$ \\
\hline
$\ell_p$-Norm Approx.  & 
\begin{tabular}{@{}l@{}}
Time: Higher than $\ell_2$ \\ 
Space: $O(k(m+n))$
\end{tabular} & 
Accuracy: 0.63--1\% drop & 
Computationally intensive \\
\hline
\textbf{ARSVD} (Proposed Method)  & 
\begin{tabular}{@{}l@{}}
Time: $O(k(m+n))
$ \\
Space: $O(k(m+n))$
\end{tabular} & 
Accuracy: 1\% increase & 
Generalization of large weight matrices \\
\hline
\end{tabular}%
}
\end{table*}

\vspace{0.1 cm}

SVDs have been used to shrink the computation and memory requirements of neural networks. However, most methods suffer from a loss of accuracy. A common method is the Truncated SVD (TSVD) for Low Rank Approximation. Here we will present the top existing methods in the literature:

Table~\ref{tab:svd_comparison} compares SVD-based compression methods in neural networks.

\begin{itemize}
\item TSVD \cite{girshick2015fast} and RSVD \cite{halko2011finding} offer efficient inference but with accuracy trade-offs
\item Transpose trick \cite{drineas2005approximating} is exact but situationally useful
\item $\ell_p$-norm \cite{candes2010power} preserves accuracy at higher cost
\end{itemize}

Key observations:
\begin{itemize}
    \item TSVD \cite{girshick2015fast} and RSVD \cite{halko2011finding} offer linear inference complexity but suffer accuracy losses
    \item The transpose trick \cite{drineas2005approximating} provides exact decomposition but has limited applicability
    \item $\ell_p$-norm methods \cite{candes2010power} preserve accuracy better at higher computational cost
\end{itemize}

\subsection{Singular Value Decomposition Limitations}

Singular Value Decomposition (SVD) is widely used and has become something of the norm for matrix factorization tasks, with significant application in dimensionality reduction and data compression. Despite its wide appeal, SVD faces several limitations that are extremely critical in the context of deep neural network compression. SVD is computationally expensive, with a time complexity of \(O(\min(m^2n, mn^2))\), which renders it inefficient for large-scale matrices commonly encountered in deep networks. To mitigate this, techniques such as \textit{randomized SVD} and \textit{truncated SVD} have been introduced \cite{9513402}. These methods reduce the computation by approximating or truncating singular values. However, this often results in a drop in accuracy and performance due to neglecting informative spectral components. For our compression task, preserving the meaningful structures present in the weight matrices is critical to maintaining performance. Although truncated or randomized approaches speed up computation, they ignore significant components in the matrix's singular value entropy distribution. Heuristically preserving features of the weight matrices is therefore a core requirement in high-performance neural network compression.

SVD is also sensitive to noise, often capturing variations that do not reflect the underlying data structure. Existing robust variants such as \textit{Proper Orthogonal Decomposition} attempt to mitigate this through minimizing reconstruction error under noisy conditions \cite{epps2019singular}. Nevertheless, handling noise effectively remains a challenge, and continues to be an active area of matrix approximation research \cite{fan2021robust}. To address the scalability of SVD in high-dimensional contexts, fast SVD variants using a block-wise or incremental strategy have been developed \cite{xu2023fast}. These methods offer trade-offs between efficiency and accuracy. Our work aims to strike a balance between these fast techniques and the accuracy retention of full SVD for real-time neural network compression.

Structured pruning techniques, which are directly complementary to matrix decompositions, adaptively reduce network size through sensitivity analysis and loss-aware heuristics. Recent work has demonstrated pruning strategies that dynamically adjust pruning rates based on connection sensitivity, and the loss landscape \cite{sakai2022structured}. These proposed methods suggest that adaptivity, guided by information-theoretic or optimization-driven metrics can yield improvements in performance. Our approach is motivated by these insights, in pairing them with spectral entropy to determine and select a rank adaptively, aiming to retain high-information components in each layer while ensuring efficiency during computation.

\section{Problem Formulation}

\noindent Consider a fully connected (FC) layer of a neural network:
\[
f_{\theta}: \mathbb{R}^d \to \mathbb{R}^C,
\]
parameterized by $\theta$.  At layer~$l$, let 
\[
W^{(l)}\in\mathbb{R}^{m\times n},\quad h^{(l)}\in\mathbb{R}^{n},\quad 
b^{(l)}\in\mathbb{R}^{m}
\]
denote the weight matrix, input activation, and bias, respectively.  The standard forward pass is
\[
h^{(l+1)} \;=\; \sigma\bigl(W^{(l)}\,h^{(l)} + b^{(l)}\bigr),
\]
where $\sigma(\cdot)$ is an non-linear activation function.

\subsection*{Cost of Full-Rank Multiplication}
A direct matrix–vector product $W^{(l)}h^{(l)}$ costs
\[
O\bigl(m\times n\bigr)
\]
operations.  When $m, n$ are large, this dominates both storage and runtime, becoming computationally unfeasible.

\subsection*{Low-Rank Decomposition}
We seek a rank‐$k$ approximation via singular value decomposition:
\[
W^{(l)} = U\,S\,V^T, 
\quad S=\mathrm{diag}(s_1,\dots,s_r), 
\quad r=\min(m,n).
\]
Here $s_1\ge s_2\ge\cdots\ge s_r\ge0$ are the ordered singular values.  A truncated representation
\[
\tilde W^{(l)} = U_k\,S_k\,V_k^T,
\]
with 
\[
U_k\in\mathbb{R}^{m\times k},\quad 
S_k=\mathrm{diag}(s_1,\dots,s_k),\quad 
V_k^T\in\mathbb{R}^{k\times n},
\]
reduces storage from $mn$ to $k(m+n)$ parameters.

\subsection*{Complexity of Low-Rank Forward Pass}
Using $\tilde W^{(l)}$ yields three successive steps:
\begin{align*}
&\text{(i) } V_k^T\,h^{(l)}\quad &&O(k\,n),\\
&\text{(ii) } S_k\,(V_k^T\,h^{(l)})\quad &&O(k),\\
&\text{(iii) } U_k\bigl[S_k(V_k^T\,h^{(l)})\bigr]\quad &&O(k\,m).
\end{align*}
Summing,
\[
\underbrace{O(k\,n)}_{\text{(i)}} + \underbrace{O(k)}_{\text{(ii)}} + \underbrace{O(k\,m)}_{\text{(iii)}}
\;=\; O\bigl(k(m+n)\bigr).
\]
Since $k\ll\min(m,n)$, this satisfies the lower bound
\[
O\bigl(k(m+n)\bigr) \;<\; O(mn).
\]

\subsection*{Entropy-Based Rank Selection}

Given the singular values \(s_1, \ldots, s_r\) of a weight matrix \(W \in \mathbb{R}^{m \times n}\), we define the normalized spectrum:
\[
p_i = \frac{s_i}{\sum_{j=1}^r s_j}, \quad i = 1, \dots, r.
\]
The total spectral entropy is:
\[
H_{\text{total}} = -\sum_{i=1}^r p_i \log p_i,
\]
and the partial entropy over the top \(k\) singular values is:
\[
H(k) = -\sum_{i=1}^k p_i \log p_i.
\]
We select the smallest \(k\) such that:
\[
H(k) \ge \tau H_{\text{total}}, \quad \tau \in (0, 1],
\]
ensuring that a fraction, at minimum \(\tau\) of the spectral information is retained. This enables the adaptive rank selection compression when the spectrum is concentrated and less when it is better distributed.

\subsection*{Adaptive Rank Singular Value Decomposition (ARSVD)}

The ARSVD procedure compresses each weight matrix as follows:

\begin{itemize}
  \item Compute the SVD: \(W = U S V^T\).
  \item Normalize \(s_i\) to compute \(p_i\) and \(H_{\text{total}}\).
  \item Select the smallest \(k\) such that \(H(k) \ge \tau H_{\text{total}}\).
  \item Truncate the decomposition: \(\tilde{W} = U_k S_k V_k^T\).
\end{itemize}

The forward pass becomes:
\[
h^{(l+1)} = \sigma(U_k S_k V_k^T h^{(l)} + b^{(l)}).
\]

This reduces the per-layer computational complexity from \(O(mn)\) to \(O(k(m + n))\), where \(k \ll \min(m, n)\), yielding significant efficiency improvements in both memory and runtime.

\begin{algorithm}
  \caption{Adaptive‐Rank SVD Compression (\textbf{ARSVD})}
  \begin{algorithmic}[1]
    \Require $W \in \mathbb{R}^{m \times n}$, entropy threshold $\tau \in (0, 1]$
    \Ensure Compressed weight matrix $\tilde{W}$, selected rank $k$
    \State $[U, S, V^T] \gets \mathrm{SVD}(W)$ 
    \State $r \gets \min(m, n)$
    \State $s_i \gets S_{ii},\quad p_i \gets s_i / \sum_{j=1}^r s_j$
    \State $H_{\text{total}} \gets -\sum_{i=1}^r p_i \log p_i$
    \For{$j = 1$ \textbf{to} $r$}
      \State $H_j \gets -\sum_{i=1}^j p_i \log p_i$
      \If{$H_j \ge \tau H_{\text{total}}$}
        \State $k \gets j$;\quad \textbf{break}
      \EndIf
    \EndFor
    \State $U_k \gets U[:, 1{:}k],\quad S_k \gets \mathrm{diag}(s_1,\dots,s_k)$
    \State $V_k^T \gets V^T[1{:}k, :]$
    \State $\tilde{W} \gets U_k S_k V_k^T$
    \State \Return $\tilde{W},\ k$
  \end{algorithmic}
\end{algorithm}

To compress an entire neural network $M$, we apply \textbf{ARSVD} to each weight matrix.  This preserves the layer‐wise representational capacity up to the chosen entropy threshold while achieving dramatic parameter and compute savings.

\begin{itemize}
  \item \emph{Global Complexity Reduction:} Summing per‐layer costs,
  \[
    \sum_{l}O\bigl(k_l(m_l + n_l)\bigr)
    \;<\;
    \sum_{l}O\bigl(m_l n_l\bigr),
  \]
  where $k_l\ll\min(m_l,n_l)$ for each layer.
  \item \emph{Space Reduction:} Parameter reduction from $\sum_l m_l n_l$ to $\sum_l k_l(m_l + n_l)$.
\end{itemize}
\vspace{0.15 cm}
Here is an end-to-end application using the proposed \textbf{ARSVD}.

\vspace{-0.2 cm}
\begin{algorithm}
  \caption{Compressed Neural Network with \textbf{ARSVD}}
  \begin{algorithmic}[1] % Line numbering starts here
    \Require Pretrained model $M$, entropy threshold $\tau$
    \Ensure Compressed model $M^*$
    \ForAll{layers $\ell$ in $M$ with weight $W^{(\ell)}\in\mathbb{R}^{m_\ell\times n_\ell}$}
      \State $(\tilde W^{(\ell)},k_\ell)\gets\textbf{ARSVD}(W^{(\ell)},\tau)$
      \State Replace $W^{(\ell)}\leftarrow\tilde W^{(\ell)}$ and log $(m_\ell,n_\ell,k_\ell)$
    \EndFor
    \State \Return $M^*$
  \end{algorithmic}
\end{algorithm}

\section{Experimental Results}
The primary purpose of the \textbf{Experimental Results} is to assess the effectiveness of the proposed Adaptive-Rank Singular Value Decomposition (ARSVD). We evaluate the performance of several pre-trained models, including ResNet18, VGG16, and InceptionV3, with standard SVD compression, and our proposed entropy-guided \textbf{ARSVD} compression.

\subsection{Parameter Reductions and F1 Scores}
The first table below shows the number of parameters required by each model across the tested datasets. ARSVD drastically reduces the number of parameters compared to its application of SVD across ResNet18, VGG16, and InceptionV3, showcasing the efficacy of the proposed approach. The difference in the F1 score is also negligible, showcasing its minor improvement in performance with fewer parameters.

\begin{table}[ht]
\centering
\label{Parametes}
\begin{tabular}{|l|c|c|c|c|}
\hline
\textbf{Model} & \textbf{MNIST} & \textbf{CIFAR-10} & \textbf{CIFAR-100} \\
\hline
\textbf{ResNet18 (SVD)} & 11.2 & 11.2 & 11.1 \\
\textbf{ResNet18 (ARSVD)} & 9.7 & 9.8 & 9.6 \\
\hline
\textbf{InceptionV3 (SVD)} & 23.9 & 24.1 & 23.8 \\
\textbf{InceptionV3 (ARSVD)} & 17.0 & 17.5 & 17.3 \\
\hline
\textbf{VGG16 (SVD)} & 134.3 & 135.0 & 135.2 \\
\textbf{VGG16 (ARSVD)} & 106.0 & 107.0 & 108.0 \\
\hline
\end{tabular}
\vspace{0.1 cm}
\caption{Parameter Count (Millions) for Each Model Across Different Datasets}
\end{table}
\vspace{-0.3 cm}
\begin{table}[ht]
\centering
\begin{tabular}{|l|c|c|c|c|}
\hline
\textbf{Model} & \textbf{MNIST} & \textbf{CIFAR-10} & \textbf{CIFAR-100} \\
\hline
\textbf{ResNet18 (SVD)} & 99.2 & 80.1 & 55.3 \\
\textbf{ResNet18 (ARSVD)} & 99.3 & 80.4 & 55.7 \\
\hline
\textbf{InceptionV3 (SVD)} & 99.3 & 93.1 & 76.5 \\
\textbf{InceptionV3 (ARSVD)} & 99.4 & 93.5 & 77.0 \\
\hline
\textbf{VGG16 (SVD)} & 99.2 & 90.8 & 68.0 \\
\textbf{VGG16 (ARSVD)} & 99.3 & 91.0 & 68.5 \\
\hline
\end{tabular}
\vspace{0.1 cm}
\caption{F1 Score for Each Model Across Different Datasets}
\end{table}

\subsubsection*{Accuracy Comparison}
This graph compares the accuracy for the different models, where we compare the pre-trained models ResNet18, VGG16, and InceptionV3 with SVD compression, to our proposed \textbf{ARSVD}.

\begin{figure}[h!]
\centering
\includegraphics[scale=0.4]{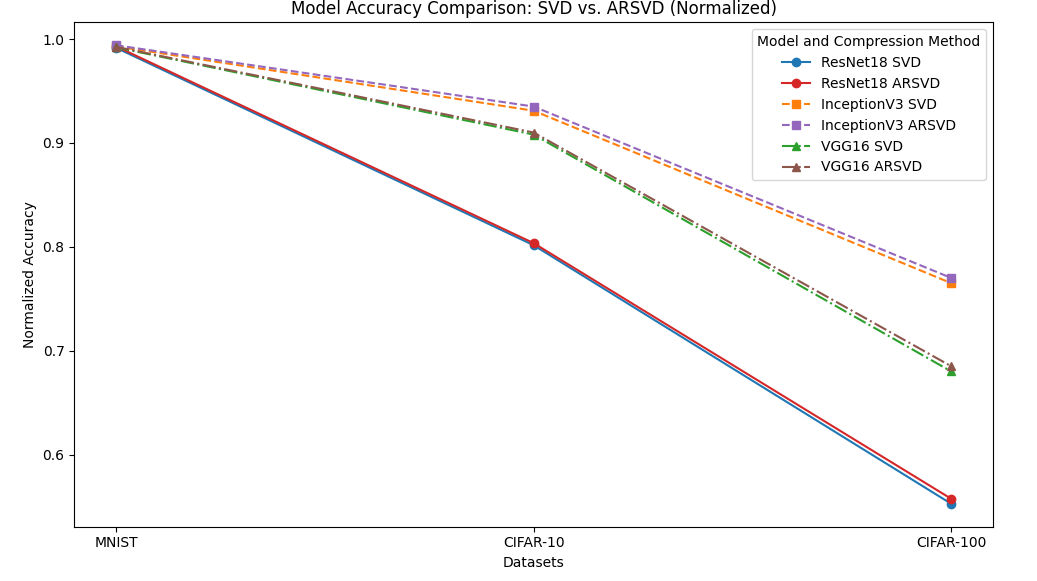}
\caption{Comparison of the accuracy of the models across MNIST, CIFAR-10, and CIFAR-100; with standard SVD compression and with \textbf{ARSVD} compression. Accuracy metrics are normalized from [0,1]}
\label{fig:accuracy}
\end{figure}

\subsubsection*{Inference Time Comparison}
This graph compares the average inference time in seconds for the different models across the 3 datasets. Where we assess the model performances with SVD compression and with our proposed \textbf{ARSVD}.

\begin{figure}[h!]
\centering
\includegraphics[scale=0.4]{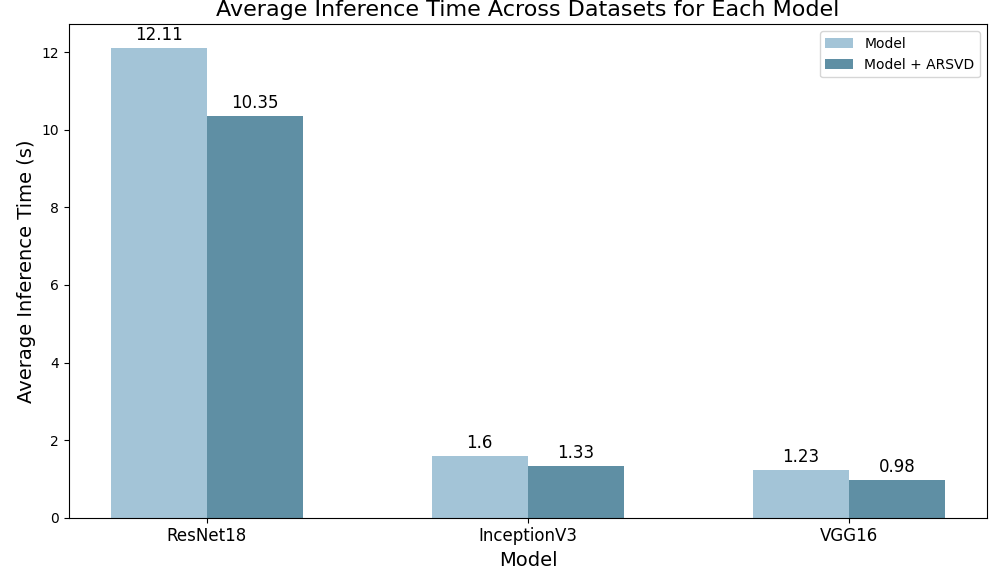}
\caption{Comparison of the average inference times (seconds) of the models across MNIST, CIFAR-10, and CIFAR-100; with SVD compared to ARSVD}
\label{fig:inference_time}
\end{figure}

\section{Discussion}
The experimental results shown above demonstrate the ability of ARSVD to effectively reduce parameter count and inference time while maintaining and slightly improving accuracy and F1 score in a variety of data sets. These improvements are particularly notable given the significant reduction in \textit{time} required for ARSVD-compressed models.
\subsection{Parameter Reduction}
As shown in Table II, our method achieves a substantial reduction in the number of parameters across all evaluated models and datasets. This compression is most notable in larger architectures like VGG16, where ARSVD reduces the parameter count by nearly 30 million on average compared to standard SVD. Such a large reduction translates directly into lower memory usage and a smaller model footprint, which is critical for deploying DNN models in computationally constrained environments.

Importantly, this reduction is not uniform across all models but adapts based on the entropy of the singular value entropy distribution in each layer. For instance, in models or layers where the spectral information is highly concentrated, ARSVD selects a smaller rank without compromising the representational capacity of the layer. This is evident in the case of ResNet18 and InceptionV3, where the reductions are consistent but proportionally smaller, reflecting a more distributed spectrum and the selective nature of the algorithm.

By tailoring the rank to the information content of each layer, ARSVD avoids the blunt truncation typical of fixed-rank SVD, allowing for improved information-preserving compression. The result is a model with significantly fewer parameters that retains and occasionally improves upon baseline classification performance.

\subsection{Accuracy Retention and Improvement}
As shown in Figure \ref{fig:accuracy}, while our method reduces model parameters it maintains and occasionally improves to its computationally expensive SVD counterpart, in both accuracy and F1 scores. This is due to ARSVD’s entropy-guided compression, which preserves only the most informative components of the weight matrices.

By discarding redundant parameters, the model avoids overfitting and focuses only on essential features, improving generalization. Additionally, the reduced space complexity with less parameters, simplifies the optimization landscape. This offers several benefits, including faster convergence and more stability during fine-tuning. These factors play a part in the slight performance improvement observed in the tested models, demonstrating that a smaller, targeted architecture can outperform larger and over-parameterized counterparts.

\subsection{Inference Time Reduction}
The decrease in parameter count directly translates to faster inference, as shown in Figure~\ref{fig:inference_time}. With fewer weights to process during forward passes, the computational overhead is significantly reduced. This makes ARSVD-compressed models more efficient for real-time or resource-constrained applications, without compromising predictive performance. The reduction is especially pronounced in deeper models like VGG16, where the original parameter count is substantially higher.

\section{Conclusion}

In this work, we introduced a novel entropy-based neural network compression method that identifies and prunes low-information parameters in deep learning models. By quantifying the contribution of each weight using entropy, our method provides a principled and adaptive approach to compression that avoids the pitfalls of arbitrary strategies, such as standard SVD. This allows the model to retain only the most informative parameters, resulting in significantly reduced memory consumption and computational complexity, with similar performance.

Through comprehensive experiments using pre-trained models such as ResNet18, InceptionV3, and VGG16 on benchmark datasets such as MNIST, CIFAR-10, and CIFAR-100, we demonstrate that our approach maintains or even improves classification performance while achieving significant compression. In particular, we observe significant improvements in inference times and parameter reductions, validating the effectiveness of using entropy as a guiding metric for model simplification. 

Overall, our results highlight the potential of using entropy in model compression as a powerful tool to build efficient and high-performing neural networks. By correlating compression with theoretical insights, our approach enables smarter pruning decisions and enables future works with more intelligent model optimization strategies. Our methods hold an exciting potential to serve as a foundation for the future of more intelligent, theory-driven compression techniques that adapt to diverse model architectures.

\section{Future Work}
While \textbf{ARSVD} demonstrates promising results in compressing neural networks with minimal performance degradation, several avenues remain open for future exploration. These include its extension to transformer architectures, such as BERT or ViT. Additionally, a theoretical study expanding upon the spectral entropy threshold, $\tau$, could provide deeper insight. Future work may investigate dynamic, layer-wise threshold tuning or treating $\tau$ as a learnable parameter to improve compression-performance trade-offs further.

\bibliographystyle{IEEEtran}
\bibliography{sample-bib}

\end{document}